\documentclass[sigconf]{acmart}

\copyrightyear{2025}
\acmYear{2025}
\setcopyright{rightsretained}
\acmConference[WSDM '25]{Proceedings of the Eighteenth ACM International Conference on Web Search and Data Mining}{March 10--14, 2025}{Hannover, Germany}
\acmBooktitle{Proceedings of the Eighteenth ACM International Conference on Web Search and Data Mining (WSDM '25), March 10--14, 2025, Hannover, Germany}\acmDOI{10.1145/3701551.3706127}
\acmISBN{979-8-4007-1329-3/25/03}

\AtBeginDocument{%
  \providecommand\BibTeX{{%
    \normalfont B\kern-0.5em{\scshape i\kern-0.25em b}\kern-0.8em\TeX}}}

\begin{document}
\title{Zero-Shot Image Moderation in Google Ads with LLM-Assisted Textual Descriptions and Cross-modal Co-embeddings}


\author{Enming Luo}
\affiliation{%
  \institution{Google Research}
}

\author{Wei Qiao}
\affiliation{%
  \institution{Google}
}

\author{Katie Warren}
\affiliation{%
  \institution{Google}
}

\author{Jingxiang Li}
\affiliation{%
  \institution{Google}
}

\author{Eric Xiao}
\affiliation{%
  \institution{Google}
}

\author{Krishna Viswanathan}
\affiliation{%
  \institution{Google Research}
}

\author{Yuan Wang}
\affiliation{%
  \institution{Google}
}

\author{Yintao Liu}
\affiliation{%
  \institution{Google}
}

\author{Jimin Li}
\affiliation{%
  \institution{Google}
}

\author{Ariel Fuxman}
\affiliation{%
  \institution{Google Research}
}

\renewcommand{\shortauthors}{Enming Luo, Wei Qiao, Katie Warren, et al.}

\begin{abstract}
We present a scalable and agile approach for ads image content moderation at Google, addressing the challenges of moderating massive volumes of ads with diverse content and evolving policies. The proposed method utilizes human-curated textual descriptions and cross-modal text-image co-embeddings to enable zero-shot classification of policy violating ads images, bypassing the need for extensive supervised training data and human labeling. By leveraging large language models (LLMs) and user expertise, the system generates and refines a comprehensive set of textual descriptions representing policy guidelines. During inference, co-embedding similarity between incoming images and the textual descriptions serves as a reliable signal for policy violation detection, enabling efficient and adaptable ads content moderation. Evaluation results demonstrate the efficacy of this framework in significantly boosting the detection of policy violating content.
\end{abstract}

\vspace{-0.1cm}

\maketitle

\vspace{-0.1cm}
\section{Introduction}

Google's advertising platform handles a massive volume of ads daily. To maintain user trust and platform integrity, every ad must be reviewed for policy violating content before it can reach the public. Ads content moderation at this scale presents a formidable challenge given the dynamic nature of online ads and evolving policy guidelines. Traditional ads content moderation approaches rely heavily on supervised machine learning models and human labeling~\cite{stretcu2023iccv} and often struggle to keep up with the volume, diversity and ever-changing nature of ads content and policy guidelines. This paper introduces a scalable and agile ads content moderation solution designed to address these challenges at Google. Our approach utilizes text-image co-embeddings to achieve zero-shot classification of policy-violating ads images. This simplifies classifier creation, making it as straightforward as describing the violation in natural language. In this paper we first detail our method for generating and validating textual descriptions that accurately capture policy guidelines. This involves using LLMs to propose candidate descriptions and then refining them through human expertise. We then discuss the use of co-embedding similarity for policy violation detection and how we use LLMs to further enhance accuracy of nuanced ads images. Finally, we present results from our implementation using the Google Ads policy enforcement platform and discuss advantages of this approach.

\section{Proposed Method}

We propose a user-centric approach where domain experts, with the assistance of LLMs, create detailed textual descriptions that encompass the policy space. These descriptions are then transformed into cross-modal co-embeddings, capturing their semantic relationship with images. During inference, incoming ads images are compared against these textual descriptions based on co-embedding similarity, facilitating efficient policy violation detection. Specifically, the method comprises three key components.


\begin{figure}[t]
  \centering
  \includegraphics[width=\linewidth]{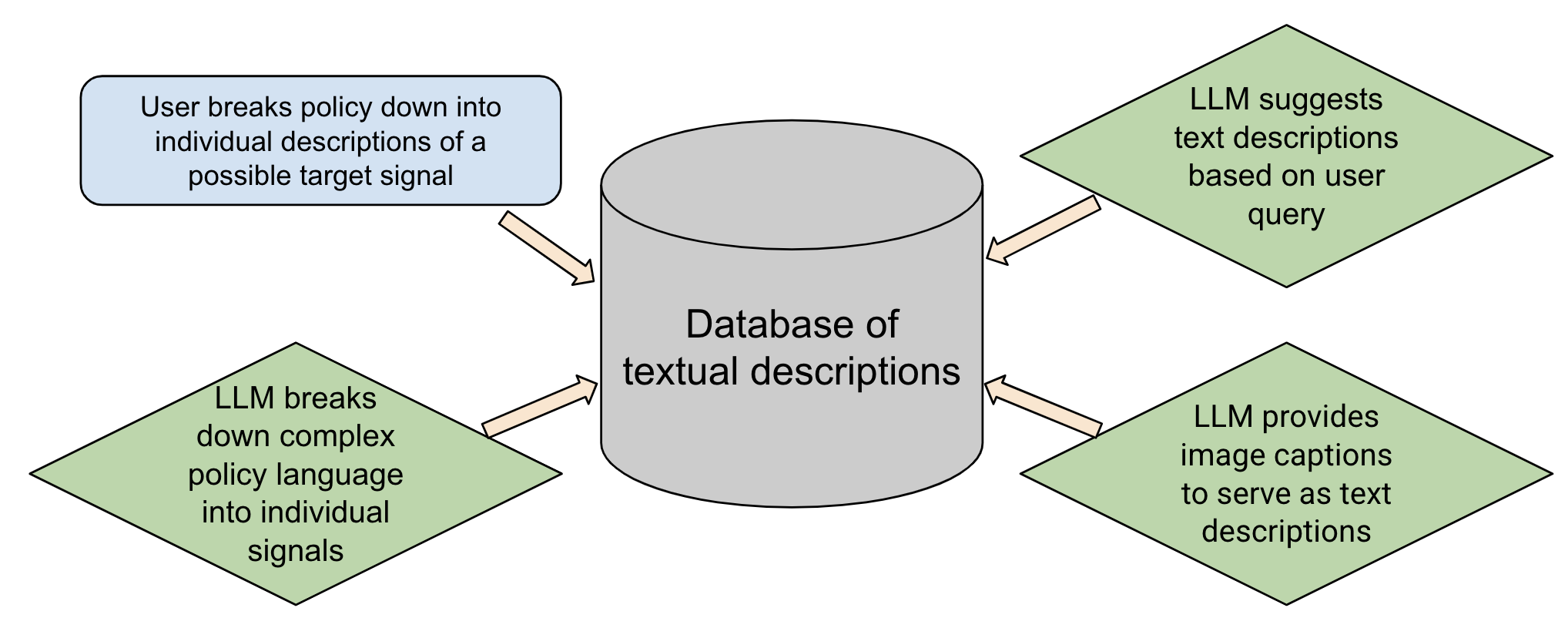}
  \caption{Generating textual descriptions for a policy by leveraging both LLMs and users with domain expertise.}
  \Description{Generating textual descriptions for a policy by leveraging both LLMs and users with domain expertise.}
  \vspace{-0.3cm}
  \label{fig:textual_descriptions}
\end{figure}

\vspace{-0.1cm}

\subsection{Generation of Textual Descriptions}
Our approach leverages LLMs to generate a rich and varied set of textual descriptions, which comprehensively captures the different modes of a given policy. We also empower users with domain expertise to craft their own textual descriptions, enhancing the LLM-generated output. Figure \ref{fig:textual_descriptions} illustrates how users can generate textual descriptions. Users can provide policy language, their own expertise, or guidance from a subject matter expert to craft descriptions. They can also leverage an LLM to uncover previously unrecorded ``blind spots'' by asking for additional suggestions. The LLM generates suggestions and assists in breaking down a complex policy into individual signals to be used as textual descriptions. Additionally, a user can provide previously human labeled violating ads images and use an LLM to caption salient signals within the images, further enriching the collection of textual descriptions.

\subsection{Validation of Textual Descriptions}
While most crafted textual descriptions are accurate, some might inadvertently misalign with the given policy. These can cause false positives, undermining the precision of the zero-shot classification model. To mitigate this, our system presents users with a selection of closely related images retrieved from existing datasets for each textual description. Leveraging their domain expertise and the visual context provided, users evaluate and label each description as either ``in-scope'' (aligned with the policy) or ``out-of-scope'' (misaligned with the policy).

To further refine the labels of the textual description set, we employ a secondary validation step. The labeled descriptions are matched against a corpus of known ads images with ground-truth labels for the given policy. In-scope descriptions that frequently match out-of-scope images are flagged as potentially problematic and subsequently removed. Similarly, out-of-scope descriptions that frequently match violating images are flagged and removed. This two-step approach ensures a high-quality set of textual descriptions, boosting the accuracy and reliability of our zero-shot classification model. Ultimately, we obtain both ``in-scope'' and ``out-of-scope'' textual descriptions for each policy.

\begin{figure}[t]
  \centering
  \includegraphics[width=\linewidth]{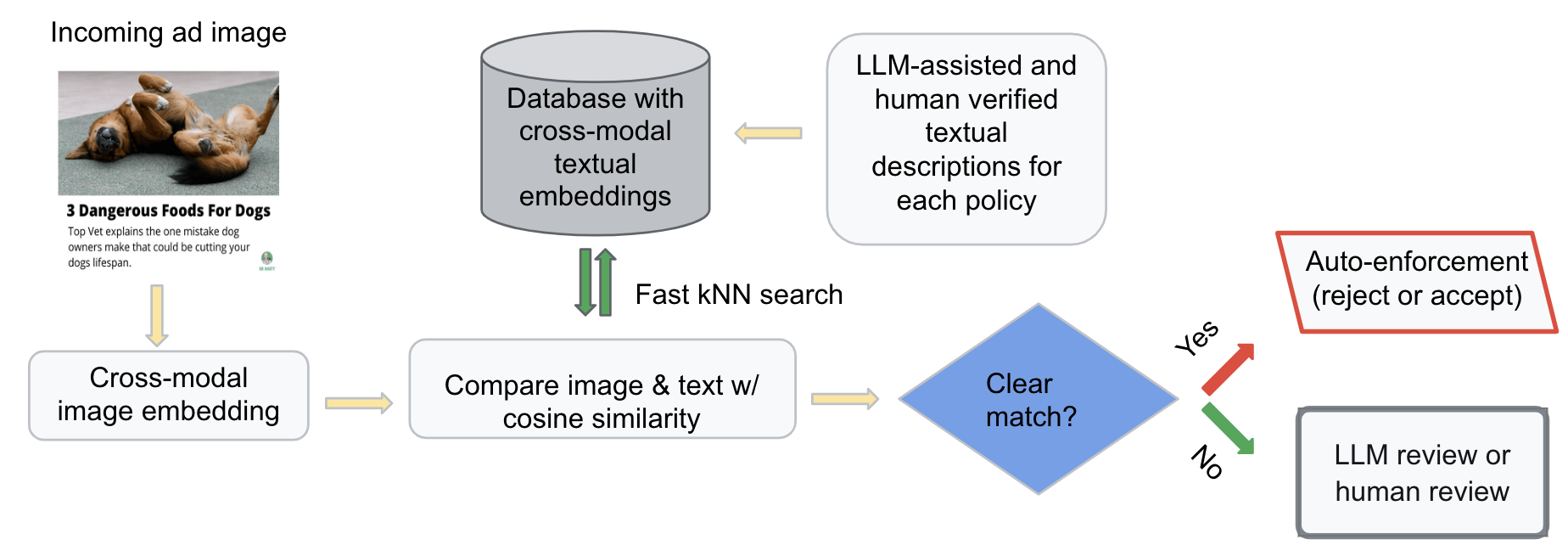}
  \caption{An end-to-end system for ads image policy enforcement, leveraging cross-modal co-embeddings and a powerful LLM.}
  \Description{An end-to-end system for ads image policy enforcement, leveraging cross-modal co-embeddings and a powerful LLM.}
   \vspace{-0.3cm}
  \label{fig:enforcement_flow}
\end{figure}

\vspace{-0.1cm}

\subsection{Policy Enforcement}

We employ a multi-stage approach leveraging cross-modal co-embeddings and the power of advanced multi-modal LLMs (illustrated in Figure \ref{fig:enforcement_flow}). 


\begin{itemize}
    \item \textbf{Cross-modal co-embedding and matching}: The labeled textual descriptions are transformed into co-embeddings. Each incoming ads image is also embedded into the same semantic space. A fast approximate kNN search~\cite{guo2020accelerating} is performed to compare the ads image with every description.
    \item \textbf{Automated decision}: If the image matches more in-scope descriptions than out-of-scope descriptions by a predetermined margin, it is automatically flagged as a policy violation. Conversely, if it matches more out-of-scope descriptions than in-scope descriptions by the same margin, it is considered policy compliant. 
    \item \textbf{LLM review}: Images that match some textual descriptions but do not fall into the clear-cut categories for automated decisions are sent to a second tier of evaluation by a fine-tuned LLM, which performs a more nuanced analysis to determine policy compliance. Our approach is used as a candidate selection mechanism, feeding ad candidates into LLM review~\cite{scalingupllmgoogleadreview} for policy decisions.
    \item \textbf{Human review}: If the fine-tuned LLM also expresses low confidence in its assessment, the image is further escalated to human expert review.
\end{itemize}
Finally, to enhance efficiency, we propagate the policy violation labels to other images that are visually similar to those already flagged.

\section{Results}
As an example, we conducted experiments on tobacco images to evaluate our approach's ability to detect tobacco-related image content. The metrics, based on human reviews, are defined as follows:
\begin{itemize}
\item \textbf{Precision}: The percentage of the true positives (TPs) among the ads labeled as positive by a model.
\item \textbf{Incremental Coverage Significance}: The percentage increase in TPs identified by a model, relative to the TPs already identified by another model.
\item \textbf{Relative recall}: The percentage of the TPs labeled by a model among all TPs labeled by all models.
\end{itemize}
The comparison between this approach and an existing binary classification model on an internal dataset is shown as below.
\begin{center} 
\begin{tabular}{ |c|c|c|c| }  
 \hline 
 & Precision & Incre. Cov. Sig. & Relative Recall \\ 
 \hline
 Our approach & \textbf{90.8\%} & \textbf{107.3\%} & \textbf{63.4\%} \\
 Baseline model &  89.1\% & 57.3\% & 48.2\% \\
 \hline 
\end{tabular} 
\end{center} 

\vspace{0.1cm}

This approach resulted in the removal of millions of policy-violating ads with high precision.


\section{Advantages of this approach}
Our approach offers several advantages for content moderation compared to other approaches:
\begin{itemize}
\item \textbf{Minimal training data is required}. The user can focus solely on designing textual descriptions. While some data is needed for the LLM and user to design textual descriptions, it doesn't need a large-scale labeled dataset for model training.
\item \textbf{Fast turnaround time} No model training is needed, and textual description design allows for faster iteration from definition to launch.
\item \textbf{Resource efficiency}. The same workflow can be used for multiple policies with one scalable search.
\end{itemize}





\bibliographystyle{ACM-Reference-Format}
\bibliography{reference}

\clearpage

\end{document}